# Square-Cut: A Segmentation Algorithm on the Basis of a Rectangle Shape

Jan Egger[1,2,3]*, Tina Kapur[1], Thomas Dukatz[2], Malgorzata Kolodziej[2], Dženan Zukić[4], Bernd Freisleben[3], Christopher Nimsky[2]

1 Department of Radiology, Surgical Planning Laboratory, Brigham and Women's Hospital, Harvard Medical School, Boston, Massachusetts, United States of America, 2 Department of Neurosurgery, University of Marburg, Marburg, Germany, 3 Department of Mathematics and Computer Science, University of Marburg, Marburg, Germany, 4 Computer Graphics Group, University of Siegen, Siegen, Germany

## Abstract

We present a rectangle-based segmentation algorithm that sets up a graph and performs a graph cut to separate an object from the background. However, graph-based algorithms distribute the graph's nodes uniformly and equidistantly on the image. Then, a smoothness term is added to force the cut to prefer a particular shape. This strategy does not allow the cut to prefer a certain structure, especially when areas of the object are indistinguishable from the background. We solve this problem by referring to a rectangle shape of the object when sampling the graph nodes, i.e., the nodes are distributed non-uniformly and non-equidistantly on the image. This strategy can be useful, when areas of the object are indistinguishable from the background. For evaluation, we focus on vertebrae images from Magnetic Resonance Imaging (MRI) datasets to support the time consuming manual slice-by-slice segmentation performed by physicians. The ground truth of the vertebrae boundaries were manually extracted by two clinical experts (neurological surgeons) with several years of experience in spine surgery and afterwards compared with the automatic segmentation results of the proposed scheme yielding an average Dice Similarity Coefficient (DSC) of 90.97±2.2%.

Funding: This work is supported by NIH P41RR19703 and R03EB013792. Its contents are solely the responsibility of the authors and do not necessarily represent the official views of the NIH. The funders had no role in study design, data collection and analysis, decision to publish, or preparation of the manuscript.

Competing Interests: The authors have declared that no competing interests exist.

* E-mail: egger@bwh.harvard.edu

## Introduction

Template-based segmentation algorithms are suitable for medical image processing, because a patient's data – mostly in the DICOM (Digital Imaging and Communications in Medicine, available: http://medical.nema.org, accessed: 2012 Jan 2) format – already offers useful information, e.g. the patient's orientation. Combined with a body landmark detection algorithm [1] that provides a landmark inside a specific organ, it is possible to choose the organ's template automatically and even get rid of a user-defined seed point inside the organ that is possibly needed by the used segmentation method.

Graph-based approaches have become quite popular during the last years. In contrast to deformable models [2] and [3] that can get stuck in local minima during the iterative segmentation (expansion) process, a graph cut algorithm provides an optimal segmentation for the constructed graph [4]. In this contribution, we present a novel graph-based algorithm for segmenting 2D objects that are rectangle shaped. The algorithm sets up a graph and performs a graph cut to separate an object from the background. However, typical graph-based segmentation algorithms distribute the nodes of the graph uniformly and equidistantly on the image. Then, a smoothness term is added [5] and [6] to force the cut to prefer a particular shape [7]. This strategy does not allow the cut to prefer a certain structure, especially when areas of the object are indistinguishable from the background. We solve this problem by referring to a rectangle shape of the object when sampling the graph nodes, i.e., the nodes are distributed non-uniformly and non-equidistantly on the image. This strategy can be useful, when areas of the object are indistinguishable from the background. To evaluate our proposal, we focus on vertebrae images from Magnetic Resonance Imaging (MRI) datasets to support the time consuming manual slice-by-slice segmentation performed by physicians – we identified an average manual segmentation time for a single vertebra of 10.75±6.65 minutes for our spine datasets. The ground truth of the vertebrae boundaries were manually extracted by two clinical experts (neurological surgeons) with several years of experience in spine surgery and afterwards compared with the automatic segmentation results of the proposed scheme yielding an average Dice Similarity Coefficient (DSC) [8] and [9] of 90.97±2.2%.

Diseases of the spine are quite common, especially due to degenerative changes of the ligamentary and ossuary structures. With increasing stenosis of the spinal cord the limitations of the patients in all-day life worsen and the current development of the population's structure leads to a growing part of older patients with a more frequent insistence for surgical treatment [10], [11] and [12]. When making the decision for adequate procedure neuro-imaging plays a main role for estimating the dimension of surgical treatment. MRI-imaging of course is particularly suitable for the assessment of spinal structures such as nerve roots, intervertebral discs and ligamentary constitution without radiation exposure. Nevertheless, certain changes of the vertebra due to osteoporosis, fractures or osteophytes require an evaluation of the bone structures via Computed Tomography (CT)-scan including radiation exposure [13] and [14]. With our series of patient datasets we try to illustrate the capability of MRI-segmentation to reconstruct the vertebral body without x-ray examination. Consequently, the numbers of pre-operative examinations can be reduced affecting radiation exposure costs and time-management.

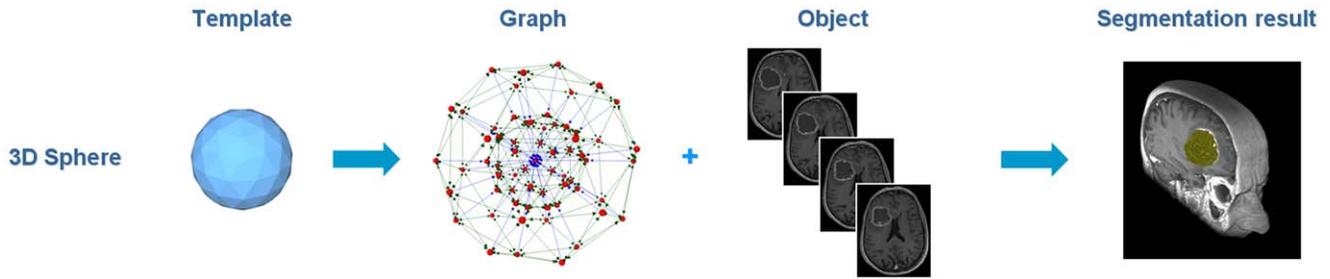

**Figure 1. Principle workflow of a segmentation scheme for Glioblastoma multiforme (GBM) in 3D.** A polyhedron (left) is used to set up a 3D graph. Then, the graph is used to segment the GBM in a Magnetic Resonance Imaging (MRI) dataset.
doi:10.1371/journal.pone.0031064.g001

For vertebrae segmentation several algorithms have been proposed in the literature. 2D segmentation approaches are mostly applied to manually identified, best suitable cross-sections [15], [16], [17] and [18]. Automatic selection of best slice was done by Peng et al. [19] and independent segmentation of the vertebral disks have been done by Michopoulou et al. [15] and Carballido-Gamio et al. [18]. Thereby, the approach from Huang et al. [18] uses normalized cut algorithm with Nyström approximation and achieves Dice Similarity Coefficients for six patients of about 93%–95%. The method from Michopoulou et al. [15] uses atlas registration of intervertebral disks, and provides DSC between 84% and 92%. The methods from Shi et al. [16] and Peng et al. [19] are both top-down approaches and Shi et al. use statistical pattern recognition for spinal cord extraction. A manually defined window is used as initialization for disk detection, and this window slides along the detected spinal cord. The authors report 96% detection rate. Peng et al. [19] do a fully automatic analysis of the whole-spine MR images. Disk clues are located by convolving a disk model with an entire MR image and a polynomial line is fit to those clues. The polynomial line has an intensity profile along which extrema indicate possible disks or vertebral bodies. It was tested on five datasets, with 100% vertebral body detection and about 95% vertebral body corner detection. Huang et al. [17] have performed the segmentation in three stages: AdaBoost-based vertebra detection, detection refinement via robust curve fitting, and vertebra segmentation by an iterative normalized cut algorithm. DSC was around 95%. This method could be called hybrid: it uses bottom-up approach for detecting vertebral body centers, but then it uses a top-down approach to segment vertebral bodies.

In contrast to the 2D approaches, 3D approaches mostly rely on user initialization. To extract the approximate spine position Yao et al. [20] use Hounsfield values and Klinder et al. [21] use CT rib cage segmentation method. The methods from Stern et al. [22], Weese et al. [23] and Hoad et al. [24] segment vertebrae independently. A very tedious initialization was used from Hoad et al., and manual corrections applied afterwards. The segmentation from Stern et al. is performed by optimizing the parameters of a 3D deterministic model of the vertebral body, aiming at the best alignment of the deterministic model and the actual vertebral body in the image. The authors estimated a 61% success rate for MRI and 84% for CT. Weese et al. use polygonal vertebra model and manual initialization. Internal energy reflects statistical shape, and external energy relies on image gradients. Method iterations consist of a surface detection step and a mesh reconfiguration step. The authors report 0.93 mm as the mean segmentation error. Top-down approaches are presented by Yao et al. [20], Ghebreab et al. [25] and Klinder et al. [21], i.e. they start from global position and approximate shape of the spine, and use that information to better fit segmentation surfaces to actual vertebrae in the images. Yao et al. focus on routine chest and abdominal CT images. The spinal canal is extracted using a watershed algorithm and directed acyclic graph search. The vertebrae are segmented by using a four-part vertebra model. The spinal column was correctly partitioned in 67 out of 69 cases. Ghebreab et al. use manual initialization for first vertebra and global spine shape. It uses B-spline surfaces with $12 \times 12$ control points for surface representation. It uses statistical spine shape for initializing segmentation of an adjacent vertebra. The mean shapes of four different lumber vertebrae are independently constructed. The method was tested on six CT images, but execution time and precision were not given. Klinder et al. initialized the global spine position by an automated rib cage segmentation method. A statistical constellation model for vertebrae is applied on a global scale to obtain an approximate position of individual vertebrae. Local adaptations of each vertebra are similar to the approach from Weese et al. The method was evaluated on ten thoracic CT datasets. The segmentation error was $1.0 \pm 0.3$ ($\mu \pm s$ mm).

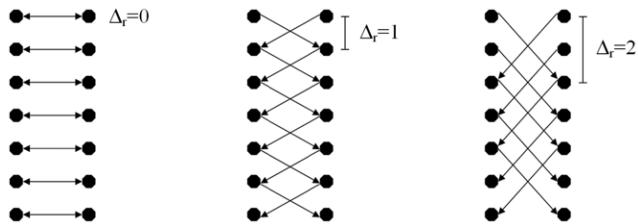

**Figure 2. Intercolumn arcs that have been constructed with different delta values: $\Delta_r = 0$ (left), $\Delta_r = 1$ (middle) and $\Delta_r = 2$ (right).**
doi:10.1371/journal.pone.0031064.g002

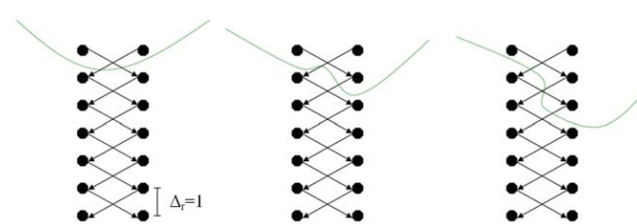

**Figure 3. Basic concept of a cut (green) of intercolumn arcs between two rays for a delta value of one ($\Delta_r = 1$).** Left and middle: same cost for a cut ($2\cdot\infty$). Right: higher cost for a cut ($4\cdot\infty$).
doi:10.1371/journal.pone.0031064.g003

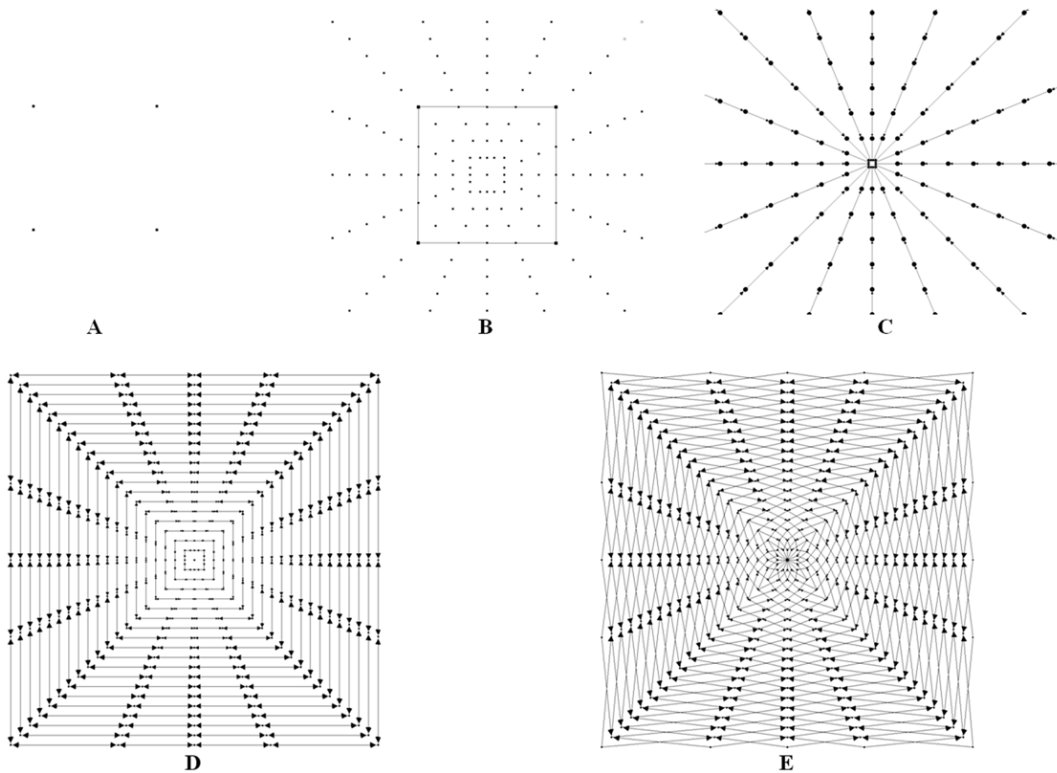

**Figure 4. The principle graph construction for a square.** A: square template defined by four corners. B: nodes set up with the square template. C: z-arcs $A_z$ along the rays. D: r-arcs $A_r$ between the rays ($\Delta_r = 0$). E: r-arcs $A_r$ between the rays ($\Delta_r = 1$).
doi:10.1371/journal.pone.0031064.g004

Some 2D methods avoid usage of computationally expensive operations and keep execution times within a few seconds [15] and [17]. Others have longer running times: forty seconds [16] and one minute [18]. Peng et al. [19] do not provide execution time. All existing 3D approaches have long running times: 1–15 minutes [22], 5–10 minutes [24], a few minutes [23] and for [21] similar to or more than [23] (not explicitly stated). Yao et al. [20] and Ghebreab et al. [25] do not provide execution time.

The paper is organized as follows. Section 2 presents the details of the proposed algorithm. Section 3 discusses the results of our experiments. Section 4 concludes the paper and outlines areas for future research.

## Methods

The proposed segmentation algorithm starts by setting up a directed graph from a user-defined seed point that is located inside the object to be segmented. The basic principle was recently developed and used by the authors for a medical software system for volumetric analysis of different cerebral pathologies – glioblastoma multiforme (GBM) [26], pituitary adenomas [27] and cerebral aneurysms [28] – in MRI datasets [29]. However, these cerebral pathologies were spherical or elliptical shaped 3D objects [30] and therefore the segmentation scheme was not appropriate for our spine datasets. For better understanding of this paper the overall principle for GBM segmentation with a sphere template is presented in Figure 1: a polyhedron (left) is used to set up a 3D graph. Then, the graph is used to segment the GBM in a Magnetic Resonance Imaging (MRI) dataset.

To set up the graph, points are sampled along rays that are sent through the contour of a square template. The sampled points are the nodes $n \in V$ of the graph $G(V, E)$ and $e \in E$ is the corresponding set of arcs. There are arcs between the nodes and arcs that connect the nodes to a source $s$ and a sink $t$ to allow the computation of a s-t cut (note: the source and the sink s, t$\in$V are virtual nodes). The arcs $<v_i, v_j> \in E$ of the graph $G$ connect two nodes $v_i, v_j$. There are two types of $\infty$-weighted arcs: z-arcs $A_z$ and r-arcs $A_r$ (Z is the number of sampled points along one ray $z = (0,\ldots,Z-1)$ and R is the number of rays sent out to the contour of an object template $r = (0,\ldots,R-1)$), where $V(x_n, y_n)$ is a neighbor of $V(x,y)$ – in other words $V(x_n, y_n)$ and $V(x,y)$ belong to two adjacent rays [31] and [32]:

$$
\begin{aligned}
A_z &= \{\langle V(x,y), V(x,y-1)\rangle | y > 0\} \\
A_r &= \{\langle V(x,y), V(x_n, \max(0, y - \Delta_r))\rangle\}
\end{aligned}
\tag{1}
$$

The arcs between two nodes along a ray $A_z$ ensure that all nodes below the contour in the graph are included to form a closed set (correspondingly, the interior of the object is separated from the exterior in the data). The arcs $A_r$ between the nodes of different rays constrain the set of possible segmentations and enforce smoothness via the parameter $\Delta_r$. The arcs for different delta values are presented in Figure 2: $\Delta_r = 0$ (left), $\Delta_r = 1$ (middle) and $\Delta_r = 2$ (right). The larger this parameter $\Delta_r$ is, the larger is the number of possible segmentations. In Figure 3 the basic concept of a cut (green) of intercolumn arcs between two rays for $\Delta_r = 1$ is presented. For the graphs on the left side and the middle the costs for a cut $(2 \bullet \infty)$ are the same. However, for a cut like shown on the right side of Figure 3 the costs are higher $(4 \bullet \infty)$.

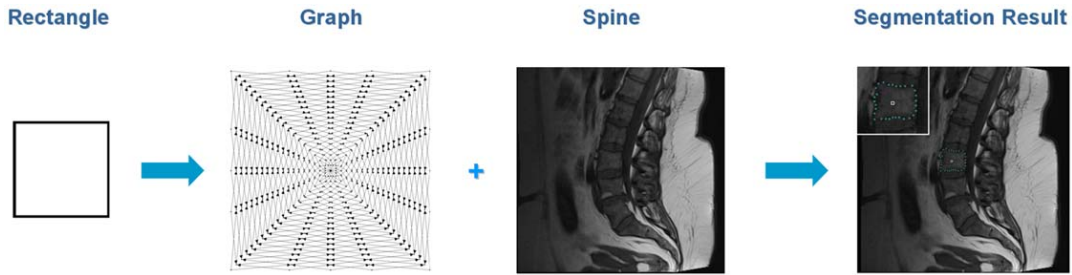

**Figure 5. Overall workflow of the segmentation algorithm.** A rectangle shape is used to set up a graph. The constructed graph is then used to segment the vertebrae in a Magnetic Resonance Imaging (MRI) scan.
doi:10.1371/journal.pone.0031064.g005

After graph construction, the minimal cost closed set on the graph is computed via a polynomial time s-t cut [33]. The s-t cut creates an optimal segmentation of the object under influence of the parameter $\Delta_r$ that controls the stiffness of the resulting contour. A delta value of zero ensures that the segmentation result has exactly the form of the predefined template (square) – and the position of the template depends on the best fit to the image's texture. The weights w(x,y) for every arc between $v \epsilon V$ and the sink or source are assigned in the following manner: weights are set to c(x,y) if z is zero; otherwise they are set to $c(x,y) - c(x,y-1)$, where $c(x,y)$ is the absolute value of the intensity difference between an average texture value of the desired object and the texture value of the pixel at position (x,y) – for a detailed description, see [34], [35], [36] and [37]. The average grey value that is needed for the calculation of the costs and the graph's weights is essential for the segmentation result. Based on the assumption that the user-defined seed point is inside the object, the average gray value can be estimated automatically. Therefore, we integrate over a small square $T$ of size $d$ centered around the user-defined seed point $(s_x, s_y)$:

$$\int_{-d/2}^{d/2} \int_{-d/2}^{d/2} T(s_x+x, s_y+y) dx dy \qquad (2)$$

**Table 1.** Comparison of manual and automatic segmentation results for nine vertebrae via the Dice Similarity Coefficient (DSC).

| No. | Volume of vertebrae (mm³) | | Number of voxels | | DSC (%) |
|---|---|---|---|---|---|
| | manual | automatic | manual | automatic | |
| 1 | 417.236 | 378.662 | 1709 | 1551 | 90.78 |
| 2 | 438.721 | 397.705 | 1797 | 1629 | 90.83 |
| 3 | 461.914 | 427.49 | 1892 | 1751 | 88.99 |
| 4 | 457.275 | 439.453 | 1873 | 1800 | 92.02 |
| 5 | 510.498 | 490.723 | 2091 | 2010 | 93.05 |
| 6 | 430.908 | 481.201 | 1765 | 1971 | 87.37 |
| 7 | 404.541 | 402.832 | 1657 | 1650 | 90.35 |
| 8 | 414.795 | 377.686 | 1699 | 1547 | 90.39 |
| 9 | 247.803 | 242.92 | 1015 | 995 | 94.93 |

doi:10.1371/journal.pone.0031064.t001

The principle of the graph construction for a square is shown in Figure 4. Image A of Figure 4 shows the square template that is used to set up the graph. Image B presents the nodes that have been sampled along the rays that have been sent through the template's surface. Note that the distances between the nodes of one ray correlate with the distances between the template's center point (or for a later segmentation, the user-defined seed point) and the template surface. In other words, for every ray we have the same number of nodes between the center point and the object's border, but the length is different. In the images C, D and E, different $\infty$-weighted arcs are shown: C: the z-arcs $A_z$ along the single rays, D: the r-arcs $A_r$ between rays with a delta value of $\Delta_r = 0$. E: same as D only with a delta value of $\Delta_r = 1$.

Setting up the nodes of the graph with the user-defined template is the most difficult step of the proposed algorithm. Generating the arcs between the nodes and the source and the sink is straightforward: there are the $\infty$-weighted arcs that depend on the geometry (intra column arcs) and the delta value (inter column arcs) used for the graph, and there are arcs that connect the nodes to the source $s$ and the sink $t$. These arcs depend on the gray values of the nodes they connect – or rather they depend on the gray value difference to an adjacent node. To integrate the user-defined template into the construction of the graph, we need the coordinates in 2D describing the object that we want to segment (e.g. for a square the corner points of the square, see Figure 4 A). Using these coordinates, the center of gravity of the object is calculated, and the object is normalized with the maximum diameter, or rather with the coordinate that has the maximum distance to the center of gravity. After the user defines a seed point in the image, the normalized object is constructed with its center of gravity point located at the user-defined seed point. Then, rays are sent out radially from the seed point through the contour of the normalized object. To calculate the intersection points of the rays with the object, the object's contour has to be closed. In our implementation, the user has to provide the object's contour as 2D

**Table 2.** Summary of results: minimum, maximum, mean $\mu$ and standard deviation $\sigma$ for manual and automatic spine segmentation.

| | Volume of vertebrae (mm³) | | Number of voxels | | DSC (%) |
|---|---|---|---|---|---|
| | manual | automatic | manual | automatic | |
| min | 247.803 | 242.92 | 1015 | 995 | 87.37 |
| max | 510.498 | 490.723 | 2091 | 2010 | 94.93 |
| $\mu \pm \sigma$ | $420.41 \pm 72.22$ | $404.3 \pm 72.98$ | 1722 | 1656 | $90.97 \pm 2.2$ |

doi:10.1371/journal.pone.0031064.t002

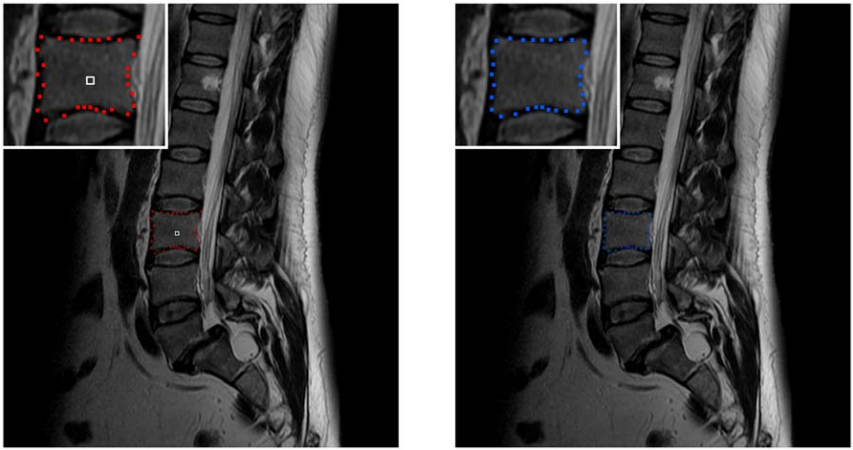

**Figure 6. Example for smoothing a vertebra segmentation result.** Left: 2D vertebra segmentation (red) of a Magnetic Resonance Imaging (MRI) dataset with a square template (number of rays = 30, number of nodes sampled per ray = 30 and delta value $\Delta_r = 4$). Right: nodes smoothed with a [0.25 0.5 0.25] kernel (one iteration).
doi:10.1371/journal.pone.0031064.g006

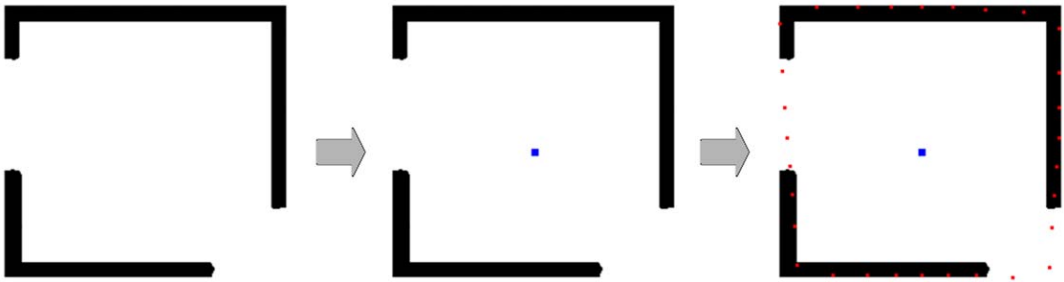

**Figure 7. Segmentation of a rectangle where parts of the border are missing.** Left: object to segment (black). Middle: user-defined seed point for the square-based segmentation (blue). Right: segmentation result (red). Note: even the missing corner in the lower right could be reconstructed.
doi:10.1371/journal.pone.0031064.g007

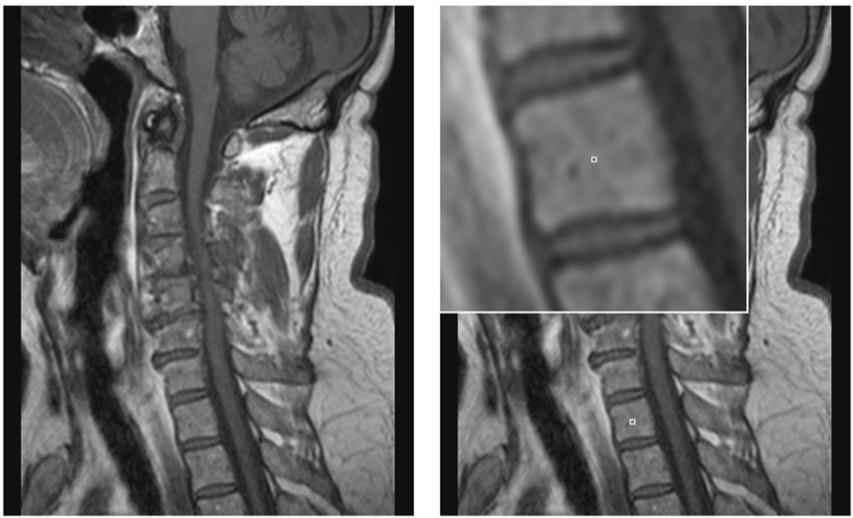

**Figure 8. Example of a spine dataset and a user-defined seed point inside a vertebra of this dataset.** Left: sagittal view of a Magnetic Resonance Imaging (MRI) spine dataset. Right: location of a user-defined seed point (white) inside a vertebra.
doi:10.1371/journal.pone.0031064.g008

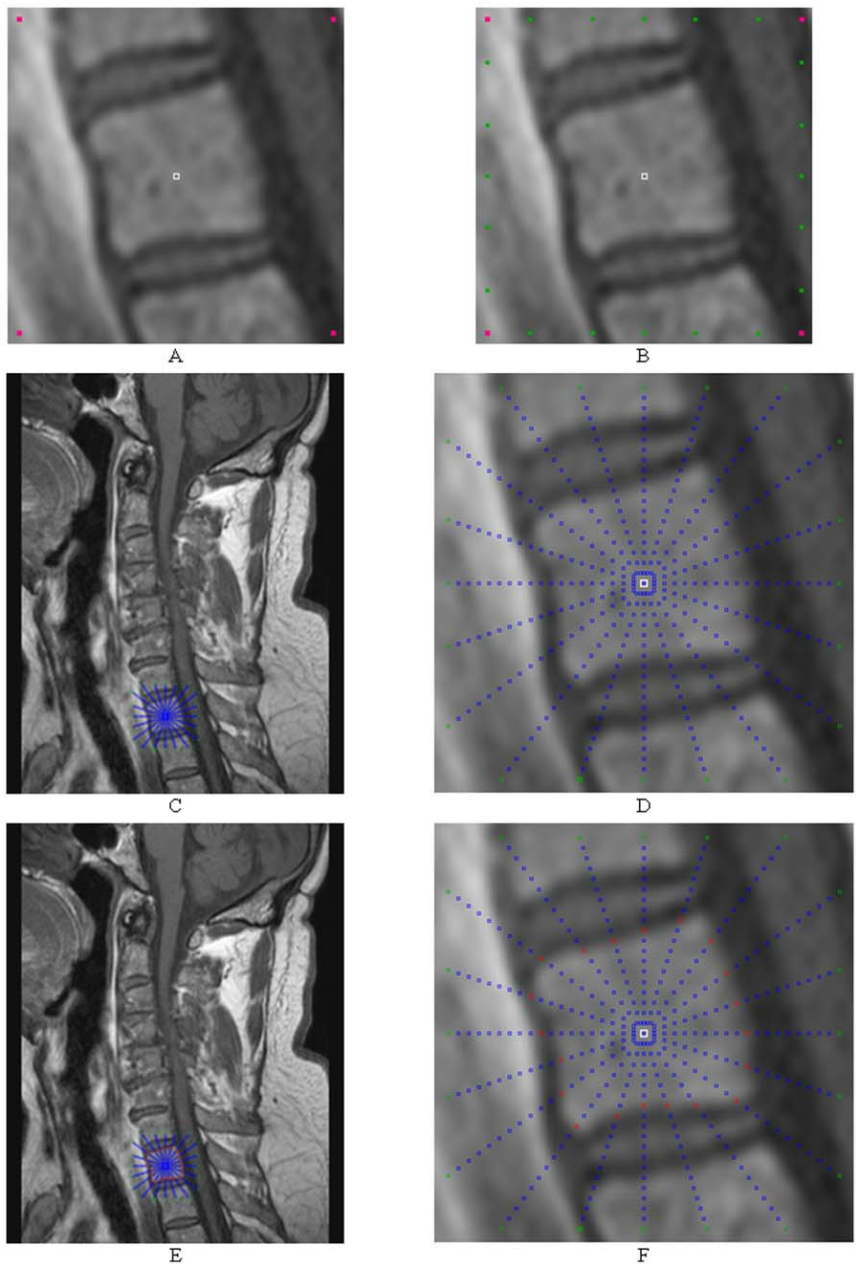

**Figure 9. Step-by-step construction of a graph and the segmentation of a vertebra.** A: seed point (white) and corners of a square template (magenta). B: intersection points where the send out rays cut the square template (green). C and D: sampled nodes for the graph (blue). E: segmentation results (red).
doi:10.1371/journal.pone.0031064.g009

coordinates ordered in clockwise direction, so we just have to connect the points one after the other and finally connect the last point with the first point to get a closed 2D contour.

The interception point of one ray with the object provides the distance between the nodes for this ray, because all rays have the same number of nodes from the center of gravity point to the intersection with the contour. For intersections that are located closer to the center of gravity point we get smaller distances, and for intersections that are located farer away from the center of gravity point we get larger distances. Calculating the intersection of a ray with a 2D object is straightforward, since it is simply a line-line intersection. One line is the actual ray and the other line is one straight line between two points of the predefined template.

## Results

To implement the presented segmentation algorithm, the MeVisLab-Platform (available: http://www.mevislab.de, accessed: 2012 Jan 2) has been used; the algorithm has been implemented in C++ as an additional MeVisLab-module. Although the foci of the prototyping platform MeVisLab are medical applications, it is possible to process images from other fields. Even when the graph was set up with a few hundred rays and hundreds of nodes where sampled along each ray, the overall segmentation (sending rays, graph construction and mincut computation) for our implementation took only a few seconds on an Intel Core i5-750 CPU, $4 \times 2.66$ GHz, 8 GB RAM, Windows XP Professional x64 Version, Version 2003, Service Pack 2.

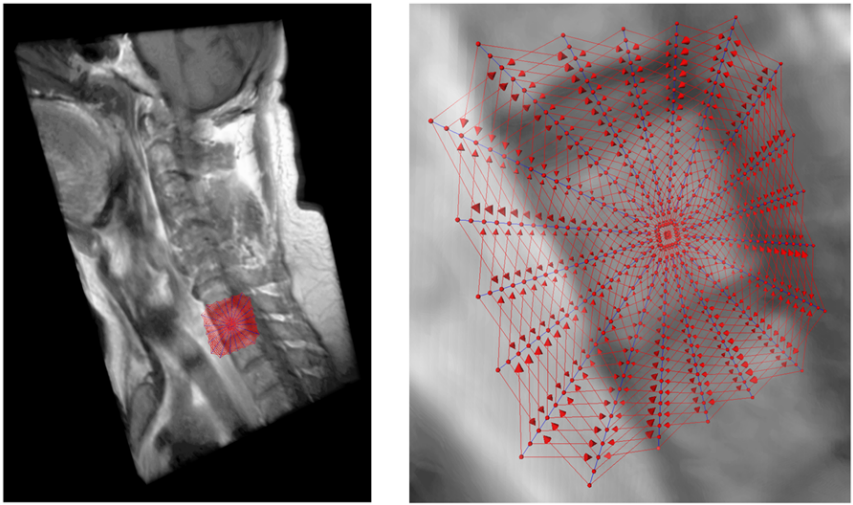

**Figure 10. 3D visualization of a Magnetic Resonance Imaging (MRI) spine dataset with a graph that has been used to segment one vertebra: intracolumn arcs (blue) and intercolumn arcs (red) with 20 rays, 20 sampled nodes per ray and a delta value of two ($\Delta_r = 2$).**
doi:10.1371/journal.pone.0031064.g010

For 2D evaluation, we used several synthetic and real images. From the clinical routine we had more than 14 datasets from over 12 patients available for testing. The overall workflow of the introduced segmentation algorithm is presented in Figure 5 (from left to right): a rectangle shape is used to set up a graph and the constructed graph is used to segment the vertebrae in a Magnetic Resonance Imaging scan.

The ground truth of the vertebrae boundaries were manually extracted by two clinical experts (neurological surgeons) with several years of experience in spine surgery and afterwards compared with the automatic segmentation results of the proposed scheme yielding an average Dice Similarity Coefficient of $90.97 \pm 2.2\%$ (Table 1 and Table 2). The Dice Similarity

Coefficient is a measure for spatial overlap of different segmentation results and is commonly used in medical imaging studies to quantify the degree of overlap between two segmented objects A and R, given by:

$$DSC = \frac{2 \cdot V(A \cap R)}{V(A) + V(R)} \qquad (3)$$

The Dice Similarity Coefficient is the relative volume overlap between A and R, where A and R are the binary masks from the automatic A and the reference R segmentation. $V(\cdot)$ is the volume (in mm$^3$) of voxels inside the binary mask, by means of counting the number of voxels, then multiplying with the voxel size. Tables 1 and Table 2 provide detailed results for several vertebrae areas of a MRI spine dataset that have been segmented with the presented algorithm. Table 1 shows the segmentation results for: volume of vertebrae (mm$^3$), number of voxels and Dice Similarity Coefficient for nine vertebrae areas. In Table 2, the summary of results: minimum, maximum, mean $\mu$ and standard deviation $\sigma$ for the nine vertebrae from Table 1 are provided. For the automatic segmentation we used the same parameter set for all vertebrae: 30 rays, 30 nodes sampled per ray and a delta value of four ($\Delta_r = 4$). The maximal length of the rays that have been sent out from the user-defined see point has been 35 mm. Furthermore we used a [0.25 0.5 0.25] kernel (one iteration) to smooth the resulting nodes that have been calculated (Figure 6).

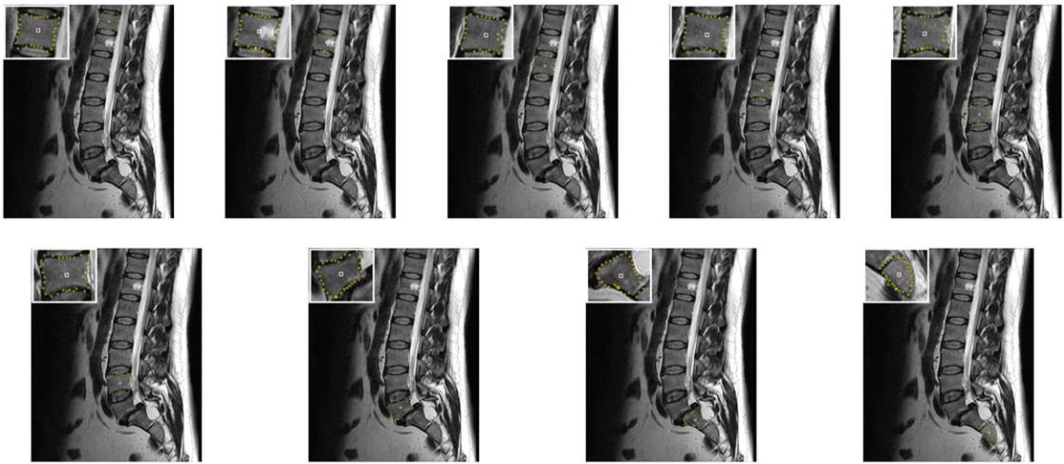

**Figure 11. 2D vertebrae segmentation (yellow) of a Magnetic Resonance Imaging (MRI) dataset with a square template (number of rays = 30, number of nodes sampled per ray = 30 and delta value $\Delta_r = 4$).**
doi:10.1371/journal.pone.0031064.g011

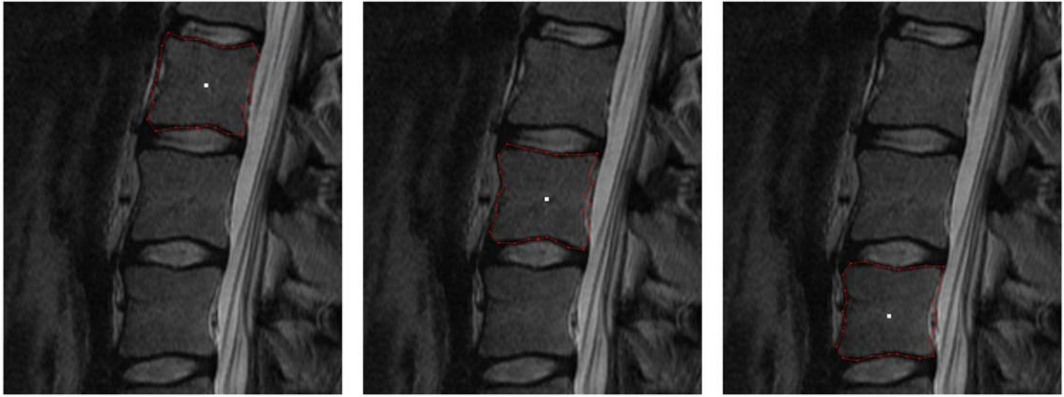

**Figure 12. 2D vertebrae segmentation (red) of a Magnetic Resonance Imaging (MRI) dataset with a square template (number of rays = 30, number of nodes sampled per ray = 30 and delta value $\Delta_r = 4$).**
doi:10.1371/journal.pone.0031064.g012

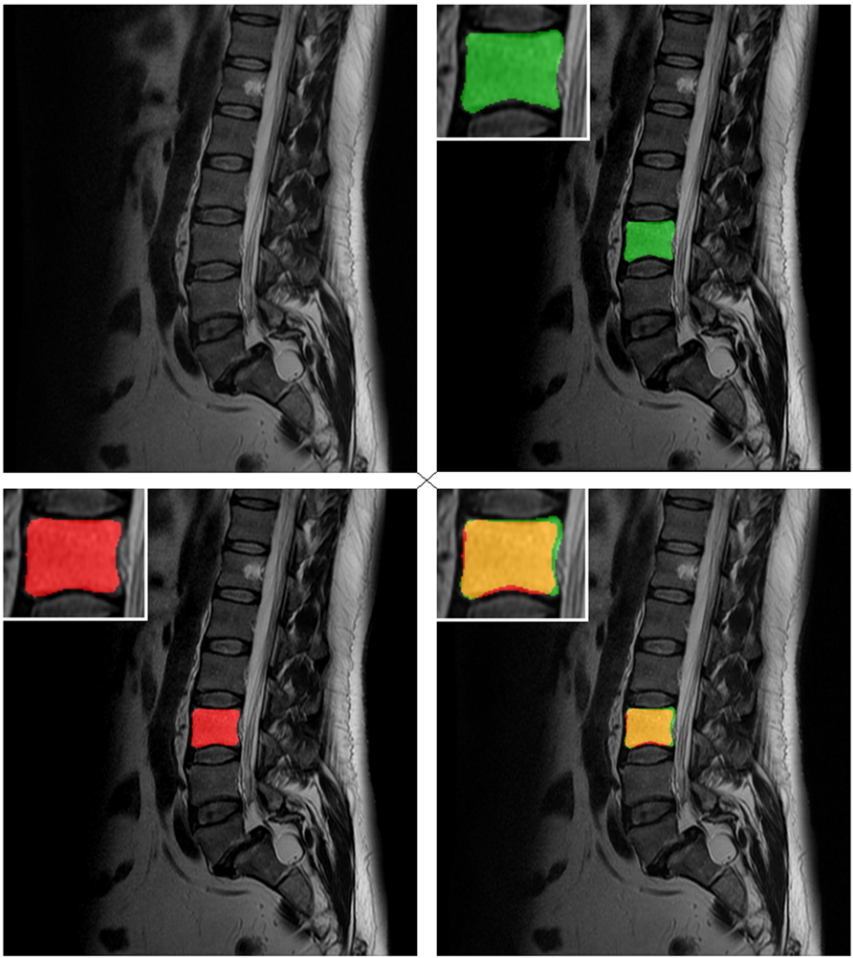

**Figure 13. Direct comparison of an automatic segmentation with a manual segmentation.** Upper right: manual segmentation mask of a vertebra (green). Lower left: automatic segmentation mask (red). Lower right: superimposed segmentation masks (manual and automatic).
doi:10.1371/journal.pone.0031064.g013

In Figure 7 the segmentation of a rectangle where parts of the border are missing is presented. On the left side of Figure 7 the object that has to be segmented (black) is shown. In the middle image the user-defined seed point (blue) for the square-based segmentation has been placed. The segmentation result (red) is shown in the rightmost image, whereby even the missing corner – in the lower right – has been reconstructed by the segmentation approach. For the segmentation we used the following parameter set: the number of rays was set to 30, the number of nodes sampled per ray was 100 and the delta value $\Delta_r$ was set to one.

Figure 8 shows on the left side a sagittal view of a MRI spine dataset. On the right side of the Figure 8 an user-defined seed point (white) has been set inside a vertebra. Figure 9 presents now step-by-step the construction of a graph and the segmentation of the vertebra of Figure 8:

> **A**: seed point (white) and corners of a square template (magenta)
>
> **B**: intersection points where the send out rays cut the square template (green)

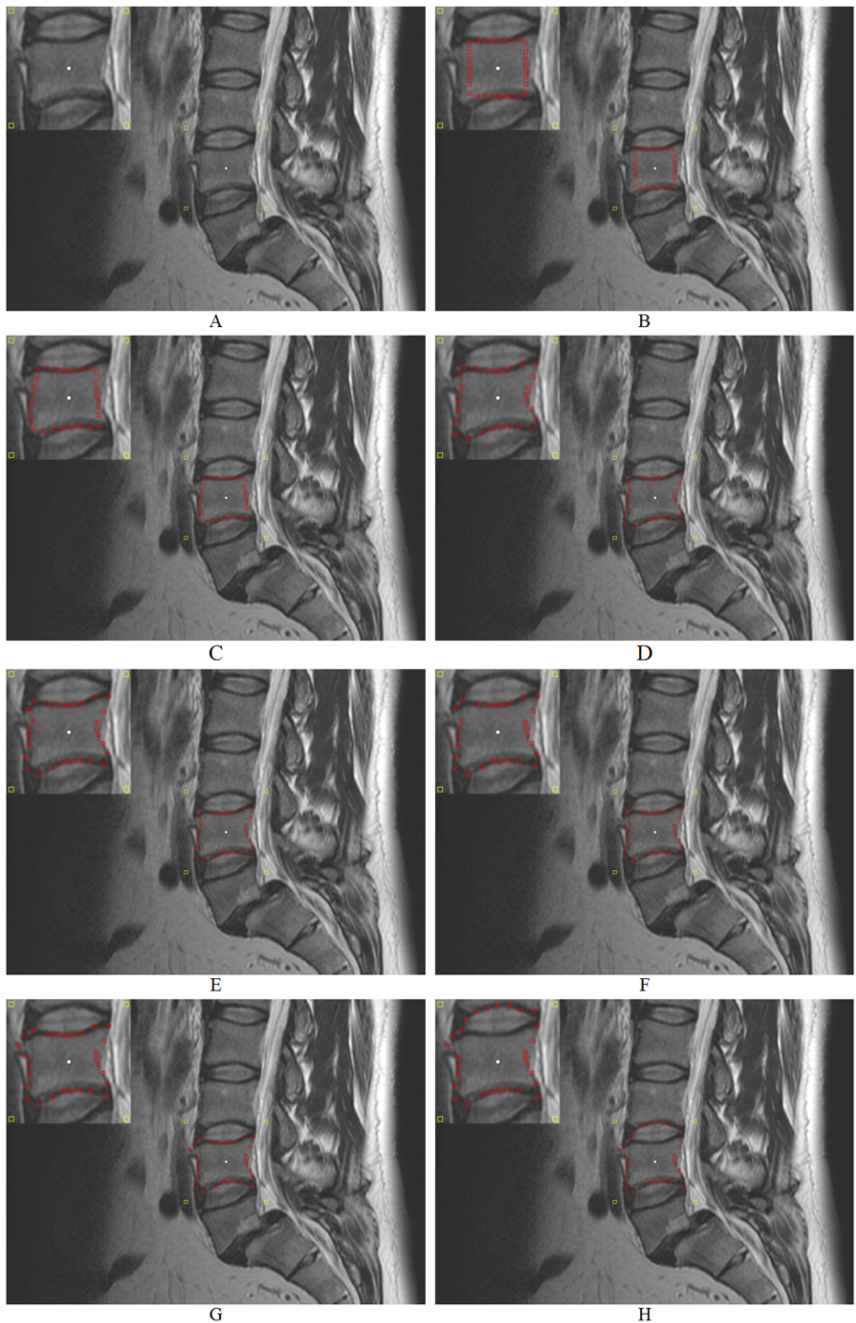

**Figure 14. Example how the ∞-weighted arcs $A_r$ (controlled via the delta value $\Delta_r$) affect the segmentation performance.** A: initial seed point (white) and corners of the square template (yellow). B–H: segmentation results (red) for different delta values $\Delta_r = 0,\dots,\Delta_r = 6$ (number of rays = 30, number of nodes sampled per ray = 40 and diameter = 40 mm).
doi:10.1371/journal.pone.0031064.g014

**C** and **D**: sampled nodes for the graph (blue)

**E**: segmentation results (red)

A 3D visualization of a MRI spine dataset with a graph that has been used to segment one vertebra is displayed in Figure 10. The intracolumn arcs of the graph are drawn in blue and the intercolumn arcs are drawn in red. The following parameter settings have been used for graph construction: 20 rays, 20 sampled nodes per ray and $\Delta_r = 2$.

The segmentation results for several vertebra of patients are shown in Figure 11 and Figure 12. The segmentations have been performed in 2D with a standard square template. Although most sides of the vertebrae are curved inwards and some vertebrae are rotated in Figure 11, the segmentation results for a square template are already reasonable. Furthermore, we have used the same parameter set for all vertebrae in Figure 11 and Figure 12, which means that the same number of rays (30), the same number of nodes sampled per ray (30) and the same delta value ($\Delta_r = 4$) for all segmentations have been used for both datasets.

Figure 13 shows the segmentation results in form of a mask for a vertebra. The original dataset is presented in the upper left of Figure 13. The manual segmentation mask of a vertebra (green) is shown in the upper right image. The lower left image presents the result of the automatic segmentation (red). Finally, the lower right image shows the superimposed manual and automatic segmentation masks.

Figure 14 shows an example how the $\infty$-weighted arcs $A_r$ (controlled via the delta value $\Delta_r$) affect the segmentation performance. Image A in Figure 14 presents the initial seed point in white inside a vertebra of a MRI spine dataset. Image A also presents the corners of the square template in yellow that has been set up with a diameter of 40 mm around the seed point. The images B-H of Figure 14 show the segmentation results in red for different delta values $\Delta_r = 0,\dots,\Delta_r = 6$ whereby the number of rays (30) and the number of nodes sampled per ray (40) have not been changed.

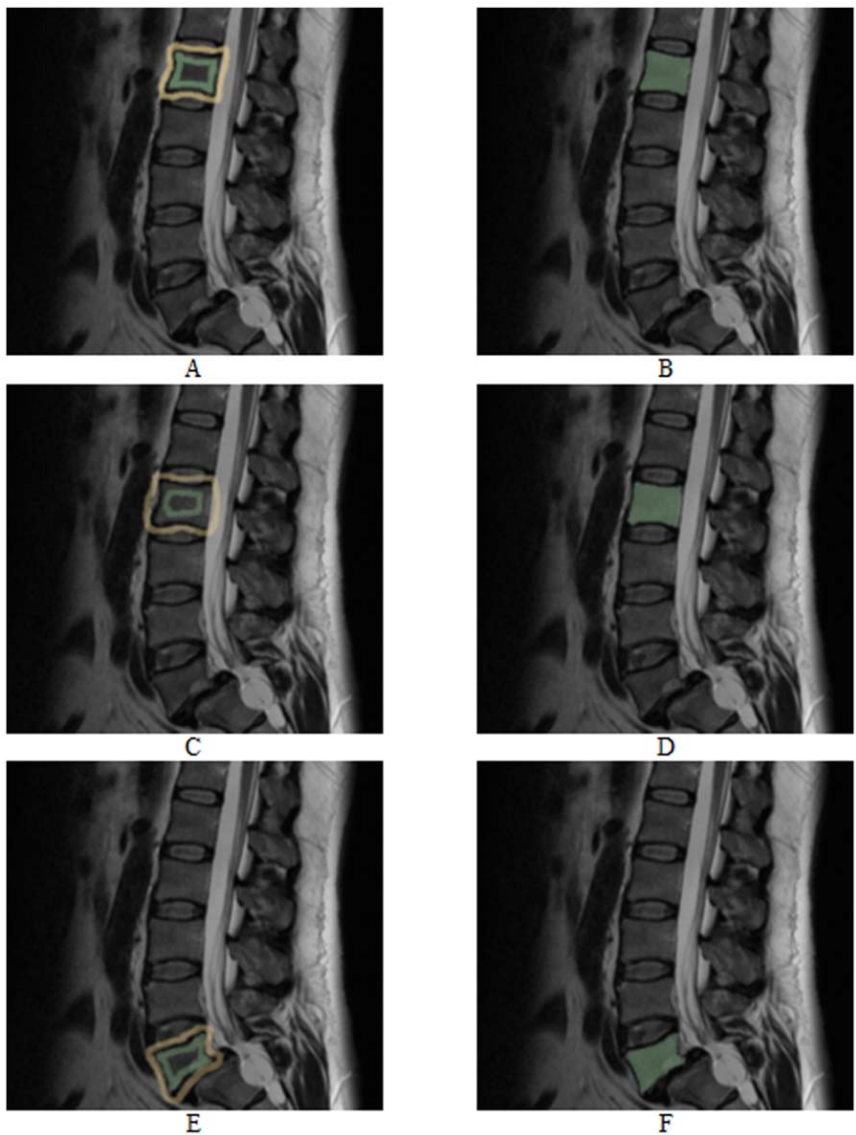

**Figure 15. Vertebrae segmentation with the *GrowCut* approach.** The images on the left side (A, C and E) show examples for a manual initialization of the algorithm: vertebra (green) and background (yellow). The images on the right side (B, D and F) present the corresponding segmentation results (green).
doi:10.1371/journal.pone.0031064.g015

In image B the delta value is zero ($\Delta_r = 0$) and therefore the resulting contour is a square, because the cut has to be on the same node level. The position of the square depends only on the gray values and edges of the image. The delta value in image C was set to one ($\Delta_r = 1$) and therefore the cut has more options and must not be on the same node level. As you can see in image C the resulting contour (red) already fits to the lower and upper border of the vertebra. However, the delta value is still too small – and therefore the possible resulting contours are to stiff – to segment the whole vertebra (see the left and right border of the vertebra). With a delta value of two ($\Delta_r = 2$) used to get the segmentation result in image D, the flexibility is high enough to segment also the left and right border of the vertebra. For the next three images E, F and G the delta values have even been increased: $\Delta_r = 3$ (E), $\Delta_r = 4$ (F) and $\Delta_r = 5$ (G). These higher delta values enables the cut to return a more ''detailed'' contour like the bulge in the upper left corner of image G. But higher delta values also increase the risk for an over-segmentation. That happened for a delta value of six ($\Delta_r = 6$) in the last image H, where the upper border of the segmentation result already returns the lower border of an adjacent vertebra.

As stated in the background paragraph, there have been published several methods – like deformable models and statistic approaches – for vertebra segmentation in the literature. All papers present detailed segmentation results and in almost all cases the computational time for their algorithms is also provided, which seem both – segmentation and time – to be similar to our results. Therefore, we decided to compare and discuss our approach with an interactive multi-label *N-D* image segmentation method called *GrowCut* from Vezhnevets and Konouchine [38]. To the best of our knowledge there has nothing been published about using *GrowCut* for spine segmentation. For testing *GrowCut* with our datasets we used an implementation that is freely available as an module for the medical platform *3DSlicer* [39] and [40]. *3DSlicer* – or *Slicer* – is a free, open source software package for visualization and image analysis primarily used in the medical domain and has been developed by the Surgical Planning Laboratory (SPL) of the Brigham and Women's Hospital in Boston. To use *GrowCut* for vertebra segmentation the user has to label a part of the vertebra and a part of the background with a simple brush tool.

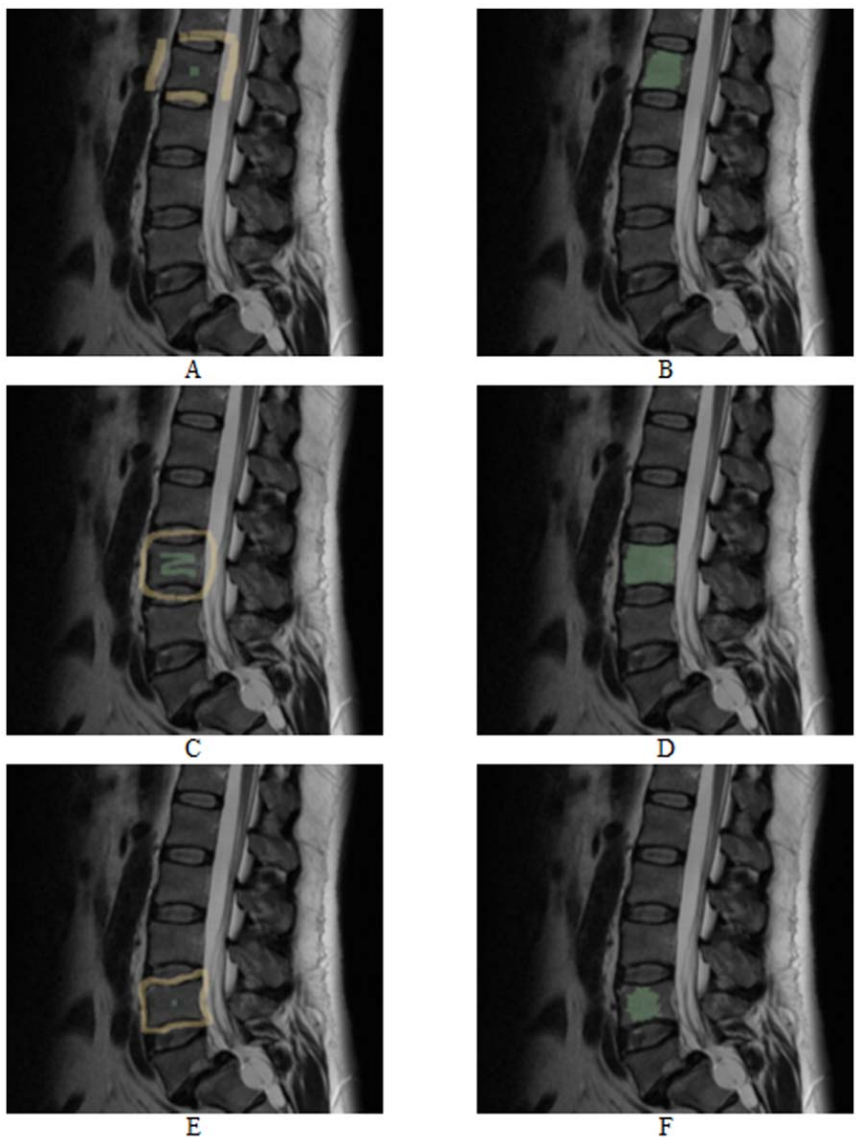

**Figure 16. As for Figure 15, vertebrae segmentation with the *GrowCut* approach.** The images on the left side (A, C and E) show examples for a manual initialization of the algorithm: vertebra (green) and background (yellow). The images on the right side (B, D and F) present the corresponding segmentation results (green).
doi:10.1371/journal.pone.0031064.g016

Figure 15 and Figure 16 present vertebrae segmentations with the *GrowCut* approach. The images on the left side (A, C and E) show examples for a manual initialization of the algorithm with the vertebrae in green and the background in yellow. The images on the right side (B, D and F) present the corresponding segmentation results in green. As you can see in Figure 15 the *GrowCut* algorithm can provide very precise results for a careful initialization. However, for a rougher initialization it can provide not satisfactory results as you can see in Figure 16 – at least for Figure B and F. We did not do an exact evaluation with the *Dice Similarity Coefficient* for the *GrowCut*, because the segmentation results depend on the user initialization. But we can already tell that for someone who knows the algorithm and knows how to deal with the initialization, the *DSC* will be around ninety percent compared with a pure manual segmentation. A big advantage of the *GrowCut* – at least for the implementation we tested – is that a user doesn't have to define any parameters. In contrast, our approach has parameters which you have to deal with, but for someone who is used to the algorithm that can be handled. A disadvantage for the *GrowCut* is the time consuming and precise initialization you sometimes need to archive good results. In contrast, our approach only needs one centered seed point inside the vertebra.

## Discussion

In this contribution, we have presented a template-based segmentation scheme for 2D objects. To the best of our knowledge, this is the first approach where the nodes of a graph-based algorithm have been arranged according to a predefined square template in a non-uniform and a non-equidistant manner on an image. Using this new type of segmentation algorithm, it is even possible to reconstruct missing corners in an object. In addition, the scaling of an object is irrelevant for the presented method. Experimental results for several 2D images based on Magnetic Resonance Imaging datasets consisting of vertebrae have indicated that the proposed algorithm requires very less computing time and gives already reasonable results even for a very simple cost function.

There are several areas of future work. For example, the cost function for the weights can be improved. Another possibility is to increase the sampling rate for the nodes near an object's border, because with an equidistant sampling rate (along the rays), there are more nodes near the user-defined seed point and less nodes going farther out. The user-defined seed point position that is located inside the object is also an issue that has to be analyzed in the future, e.g. for some images the seed point has to be chosen very carefully. In general, the presented approach provides better results if the seed point is located closer to the center of the vertebra and our method will fail or perform bad if the seed point is located very close to the border of the vertebra. One option to improve the presented algorithm is performing the whole segmentation iteratively: after segmentation has been performed, the center of gravity of the segmentation can be used as a new seed point for a new segmentation and so on. This might lead to more robustness with respect to the initialization. Furthermore, we plan to integrate our manual refinement method that takes advantage of the basic design of graph-based image segmentation algorithms [41] and [42]. Moreover, we want to enhance our segmentation algorithm to 3D. Possible is a cube template like shown in Figure 17.

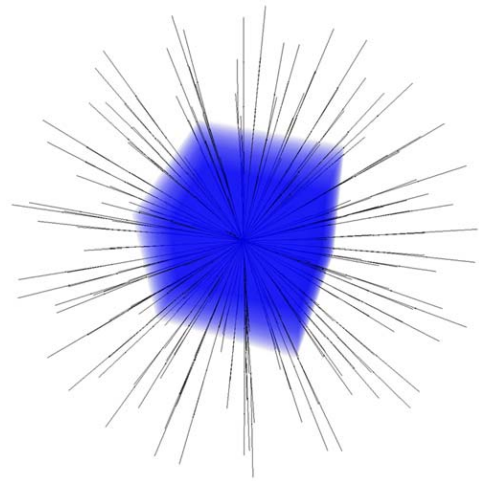

**Figure 17. Principle enhancement of the introduced 2D segmentation algorithm with a cube shape to a 3D segmentation algorithm (Cube-Cut).**
doi:10.1371/journal.pone.0031064.g017

## Acknowledgments

First of all, the authors want to thank all reviewers for their thoughtful comments. The authors would also like to thank the members of the *Slicer Community* for their contributions and moreover Harini Veeraraghavan and Jim Miller from *GE* for developing the *GrowCut* segmentation module under *Slicer*. Moreover, the authors would like to thank the physicians Dr. med. Barbara Carl and Thomas Dukatz from the neurosurgery department of the university hospital of Marburg for performing the manual slice-by-slice segmentations of the spine images and therefore providing the ground truth for the evaluation. Furthermore, the authors would like to thank Fraunhofer MeVis in Bremen, Germany, for their collaboration and especially Horst K. Hahn for his support.

## Author Contributions

Conceived and designed the experiments: JE. Performed the experiments: JE. Analyzed the data: JE TD MK. Contributed reagents/materials/analysis tools: CN. Wrote the paper: JE TK TD DZ BF.